\documentclass[11pt]{article}
\usepackage{arxiv}
\pdfoutput=1
\usepackage{cite}
\usepackage{amsmath,amssymb,amsfonts}
\usepackage{subcaption}
\usepackage{graphicx}
\usepackage{textcomp}
\usepackage{xcolor}
\usepackage{microtype}
\usepackage{mathtools}
\usepackage{cleveref}
\usepackage{changes}
\usepackage{booktabs,tabularx,array}
\newcolumntype{L}{>{\raggedright\arraybackslash}X}
\newcolumntype{C}{>{\centering\arraybackslash}X}

\newif\ifdraft \drafttrue 
\ifdraft
  \newcommand{\fixme}[1]{\todo[inline]{#1}}
\else
  \usepackage[disable]{todonotes}
  \newcommand{\fixme}[1]{}
\fi

\newcommand{\Performance}{\mathsf{Performance}}

\newcommand{\InferenceTime}{\mathsf{InferenceTime}}
\newcommand{\Memory}{\mathsf{Memory}}
\newcommand{\Energy}{\mathsf{Energy}}
\numberwithin{equation}{section}

\usepackage{float}
\usepackage{comment}
\usepackage{multirow}
\usepackage{tikz}
\usetikzlibrary{positioning,arrows.meta,fit,backgrounds}
\usepackage{algpseudocode}
\usepackage{amsmath}
\usepackage[switch]{lineno}

\usepackage{fancyhdr}
\def\BibTeX{{\rm B\kern-.05em{\sc i\kern-.025em b}\kern-.08em
    T\kern-.1667em\lower.7ex\hbox{E}\kern-.125emX}}
\begin{document}

\setlength{\abovedisplayskip}{3pt}
\setlength{\belowdisplayskip}{3pt}

\title{A Generalized Optimization Engine (GOE) for Edge AI Inference Acceleration}

\author{%
\begin{tabular}[t]{c}\textbf{Venkat R. Dasari}\\ DEVCOM Army Research Laboratory\\ Aberdeen Proving Ground, MD, USA\end{tabular}%
\hspace{2.2em}%
\begin{tabular}[t]{c}\textbf{Jakob A. Adams}\\ DEVCOM Army Research Laboratory\\ Aberdeen Proving Ground, MD, USA\end{tabular}%
\\[1.2em]
\begin{tabular}[t]{c}\textbf{Vinod K. Mishra}\\ DEVCOM Army Research Laboratory\\ Aberdeen Proving Ground, MD, USA\end{tabular}%
\hspace{2.2em}%
\begin{tabular}[t]{c}\textbf{Brian Jalaian}\\ University of West Florida\\ Pensacola, FL, USA\end{tabular}%
}
\date{\today}
\maketitle
\begin{abstract}

Artificial intelligence (AI) models have demonstrated remarkable capabilities across various domains, yet their widespread deployment is impeded by significant computational costs, particularly on resource-constrained devices. This paper explores the theoretical underpinnings of various AI model optimization techniques, algorithms, and abstractions, discussing their potential to reduce computational complexity, memory footprint, latency, and power consumption. Furthermore, we propose a comprehensive hardware (HW) and model-agnostic generalized optimization architecture that integrates these techniques for improved efficiency. Our study underscores the critical role of such a generalized optimization system in preparing model deployment over resource-constrained heterogeneous hardware in a tactical environment. As a concrete demonstration, we show that GOE-compressed language models deploy and run on a GPU-less edge CPU, and that the choice of compression method, not merely its nominal bit-width, determines whether task accuracy survives deployment.

\end{abstract}
\textit{}
\vspace{-1mm}
\section{Introduction}

Recent developments in AI models have transformed many fields, such as computer vision, autonomous navigation, military applications, and healthcare. However, AI models, particularly transformer based large language models (LLMs), are computationally complex and difficult to deploy over edge computing platforms\cite{yousri2023big}. Tactical edge is dynamic, heterogeneous and resource constrained. Real-time sensing and referencing in support of decision making is critical for mission critical operations in tactical environments\cite{coito2021intelligent}.


Creating a generalized optimization approach to explore new trade-off solutions adapted to diverse problem domains like model diversity, HW heterogeneity, and varying network conditions is quite nontrivial. Cross-platform compatibility issues like specialized HW components, HW-specific libraries and AI frameworks, and HW-specific optimization techniques create a challenge for the development of generalized approach\cite{morabito2023edge}. Although significant advances have been made in AI model optimization research, creating a comprehensive optimization solution remains an open challenge due to the complexity of the problem space.

Several approaches like model compression, pruning and efficient neural architecture search were proposed to reduce computational complexity and demand for increased resources in order to fit them on edge computing platforms with limited resources\cite{sander2025accelerating}. While these approaches are effective in model compression and inference acceleration many of them are impacted by accuracy decay diminishing their role. Custom optimization approaches that preserve the accuracy of post-optimized models are needed. Another drawback of these optimization approaches is that they are effective in a narrow context like model specific or platform specific and suffer in generalization.

\subsection{Challenges for Edge AI}
Several factors affect the performance of AI models at the tactical edge. Resource constraints and a dynamic nature affects the performance of AI models deployed over both ground and aerial autonomous systems adversely. It even contains heterogeneous computing platforms with varying degrees of computing and memory resources with variable energy dependencies. 

\section{Related Research}

Human brain inspired neural network models are widely used across a variety of fields including tactical environments. Their ability to generalize has accelerated their adoption. However, due to their computational complexity and high resource demand, their deployment over resource-limited HW is limited without optimizing them for the target HW. Recent advances in CNN and LLM optimization focus on techniques that improve efficiency, reduce computational cost, and improve scalability. Some recent work on these problems is described here. Bouzar-Benlabiod et al. (2021) give a broader overview of deep neural network architecture optimization techniques such as weight pruning, knowledge distillation, and network quantization, all of which aim to reduce model size and complexity while maintaining acceptable accuracy\cite{bouzar2021optimizing}.
Busia et al. (2022) introduce target-aware neural architecture search tool that can compose automated target-aware optimization of CNNs\cite{busia2022target}. Someki et al. (2022) presented ESPnet-ONNX, a framework that enables the deployment of deep learning models trained in research environments on various HW platforms\cite{someki2022espnet}. It addresses the challenge of cross-compatibility between research frameworks and production environments, allowing researchers to develop models that can be easily integrated. Aramdhan et al. (2023) investigate pruning methods to optimize deep neural networks used in road detection tasks \cite{aramdhan2023optimization}. Their approach is particularly beneficial for deploying deep learning models on resource-limited devices, such as autonomous vehicles requiring real-time road detection. Zhang et al. (2019) propose a compression technique for deep reinforcement learning (DRL) models by removing redundant connections and weights within the neural network, leading to a smaller model size\cite{zhang2019accelerating}.

Transformer-based LLMs are computationally more complex than CNNs. Many such model optimization techniques used for CNNs can also be used for LLM optimization for inference acceleration. However, many new optimization techniques have been developed that exclusively target LLMs. Elouargui et al. (2023) provide a comprehensive overview of techniques for making transformers more efficient\cite{elouargui2023comprehensive}. This survey explores many of them for reducing their complexity, e.g., sparse attention (focusing on a subset of relevant elements) and low-rank approximations. Liu et al. (2022) proposed dynamic sparse attention, that dynamically determines the level of sparsity within the attention mechanism. It achieves significant speedups while maintaining accuracy compared to standard attention\cite{liu2022dynamic}. The attention mechanism is a key part of transformer and LLM architectures leading to heavy computations. Traditional attention mechanisms have quadratic complexity, in which the computational cost grows quadratically with the input size. Zhuoran et al. (2021) introduced a novel attention mechanism with linear complexity, leading to substantial efficiency gains, particularly for large transformers\cite{shen2021efficient}. There are many SOTA optimization approaches, each excelling in specific contexts. However, a domain-agnostic unified optimization approach that scales across heterogeneous AI models and HW architectures remains to be developed. A consistent approach leading to predictable improvements of various models on target platforms will indicate the success of the project. 

\section{Problem Formulation}

A generalized AI model optimization approach that is agnostic to the model architecture and the target hardware is expressed mathematically, integrating detailed constraints for inference time, memory usage, energy consumption, and computational power usage. This framework provides a robust approach to optimizing neural network models for efficient deployment in resource-constrained environments. We cast this problem as constrained optimization over accuracy, latency, memory, and energy budgets (collectively called \textit{knobs} in this paper) for a given target device. See Table \ref{tab:notation} for a list of variables used in our problem formulation.

A single-objective formulation of our problem is as follows:

\begin{equation}
\label{eq:single-objective}
\begin{alignedat}{2}
\max_{m,q,p,h}\;& \Performance(m,q,p,h) \\
\text{s.t.}\;& g(m,q,p,h) \le \InferenceTime_{\text{budget}}, \\
             & h_m(m,q,p,h) \le \Memory_{\text{budget}}, \\
             & i(m,q,p,h) \le \Energy_{\text{budget}}.
\end{alignedat}
\end{equation}

For multi-objective search, we optimize accuracy \emph{and} compression subject to the same budgets:

\begin{equation}
\label{eq:multi-objective}
\begin{alignedat}{2}
\max_{m,q,p,h}\;& 
   \bigl\{\Performance(m,q,p,h),\; \mathsf{CompressionRate}(m,q,p,h)\bigr\} \\
\text{s.t.}\;& g(m,q,p,h) \le \InferenceTime_{\text{budget}}, \\
             & i(m,q,p,h) \le \Energy_{\text{budget}}.
\end{alignedat}
\end{equation}

For tractability on large models, we often restrict the search to a smaller set of knobs:

\begin{equation}
\label{eq:simplified}
\begin{alignedat}{2}
\max_{m,\rho,q,\delta,\eta}\;& \Performance(m,\rho,q,\delta,\eta) \\
\text{s.t.}\;& 0 \le \rho \le \rho_{\max}, \\
             & q \in \{4,8,16\}, \\
             & \delta \in \{0,1\}, \\
             & \eta \in \{\eta_{1},\ldots,\eta_{K}\}.
\end{alignedat}
\end{equation}

\begin{table}[t]
\centering
\caption{Notation and decision variables used in \cref{eq:single-objective,eq:multi-objective,eq:simplified}.}
\label{tab:notation}
\begin{tabularx}{\linewidth}{@{} l L @{}}
\toprule
\textbf{Symbol} & \textbf{Meaning} \\
\midrule

$m$ & Model from the repository (e.g., CNN variant, LLM family member). \\
$q$ & Quantization bit-width / policy (e.g., 4-, 8-, 16-bit; potentially mixed-precision). \\
$p$ & Pruning vector describing method and sparsity level(s). \\
$h$ & Training hyperparameters (e.g., learning rate) relevant for fine-tuning/QAT. \\
$\rho$ & Global pruning fraction (structured/unstructured). \\
$\delta$ & Distillation indicator: $0$ = none, $1$ = apply KD/self-distillation. \\
$\eta$ & Learning-rate multiplier (grid over $\{\eta_k\}$). \\
$\InferenceTime_{\text{budget}}$ & Maximum allowable inference latency on target HW. \\
$\Memory_{\text{budget}}$ & Maximum allowable memory footprint on target HW. \\
$\Energy_{\text{budget}}$ & Maximum allowable energy consumption on target HW. \\
$\Performance$ & The score of the model, such as accuracy, mAP, or perplexity.\\
\bottomrule
\end{tabularx}
\vspace{-0.5em}
\end{table}

\section{Technical Approach}

\begin{figure}[ht!]
    \centering
    \begin{tikzpicture}[
      font=\scriptsize, >=Stealth, node distance=2.6mm,
      io/.style={draw, rounded corners, fill=gray!12, align=center, inner sep=3pt, font=\scriptsize\bfseries, minimum width=3cm},
      stg/.style={draw, rounded corners, align=center, inner sep=3pt, text width=6.4cm},
      ar/.style={->, thick}]
      \node[io] (in) {CNNs \& LLMs};
      \node[stg, below=of in] (ma) {\textbf{Model Analyzer}};
      \node[stg, below=of ma] (mc) {\textbf{Model Compression}\\[1pt] NAS $\cdot$ Tensor Decomposition $\cdot$ Structured Pruning\\ Knowledge Distillation $\cdot$ Emerging Techniques};
      \node[stg, below=of mc] (hw) {\textbf{Hardware-Targeted Compression}\\[1pt] Quantization $\cdot$ Compilers (TVM, TensorRT, OpenVINO)};
      \node[io, below=5mm of hw] (out) {Target Edge Devices};
      \draw[ar] (in)--(ma); \draw[ar] (ma)--(mc); \draw[ar] (mc)--(hw); \draw[ar] (hw)--(out);
      \begin{scope}[on background layer]
        \node[draw, thick, rounded corners, fit=(ma)(mc)(hw), inner sep=7pt,
              label={[font=\scriptsize\bfseries]above:GOE Engine}] (eng) {};
      \end{scope}
    \end{tikzpicture}
    \caption{The GOE optimization engine: an input model is profiled by the Model Analyzer, compressed by architecture-aware operators, and compiled to a hardware target for deployment on edge devices.}
    \label{fig:goe-overview}
\end{figure}
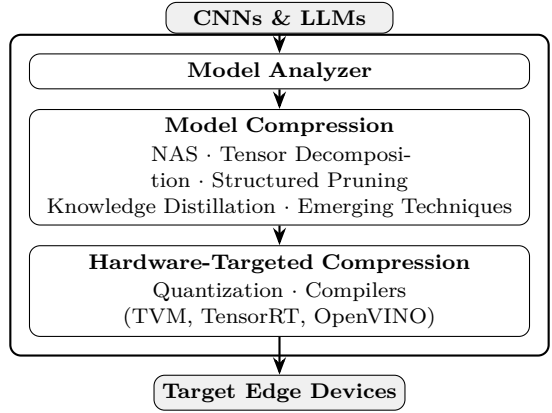

The goal of model optimization is to reduce the computational complexity, model size, and to accelerate model inference speed over resource constrained edge computing platforms. A combination of model compression approaches and compilers are leveraged to build an automated model optimization pipeline. We have surveyed current advances in AI model optimization approaches to compress models and accelerate their inference on resource constrained edge platforms. We borrow the best optimization approaches from SOTA and customized and extended them to fit them in an automated end-to-end optimization pipeline driven by a Web UI. An overview of the GOE pipeline can be found in Fig \ref{fig:goe-overview}. By building on modern optimization methods and shaping them around a focused end-to-end pipeline, the GOE framework delivers strong performance without constant hand-tuning. It cuts down on manual effort, keeps the computations efficient, and scales smoothly across different environments, making it a solid choice for complex optimization work. Next sub-sections we describe the key modules and libraries that are central to our automated generalized AI model optimization engine.



\subsection{Model Analyzer}
The gateway to our pipeline is the Model Analyzer. This component characterizes the input model to determine the model type so that the proper optimization methods can be selected to apply to it. The Model Analyzer also generates an analysis report of the model, details information such as the number of layers, number of parameters, base sparsity level, and data types, among others. Based on the information generated by the Model Analyzer and input into the pipeline, the model is routed to one or more model compression modules.

\subsection{Pruning and Quantization}
Pruning and quantization are well-established optimization techniques for compressing AI models, often leading to significant inference acceleration on resource-constrained platforms. Pruning, however, is sensitive to model architecture; it can negatively impact accuracy and typically requires fine-tuning to recover performance. Quantization reduces computational complexity and model size by lowering precision, but it can introduce quantization errors that degrade accuracy. Therefore, a robust strategy for accuracy recovery is essential, and the target hardware often dictates the specific precision levels to which a model can be effectively quantized.

\subsubsection{Optimal Brain Compression}
We have adopted Optimal brain compression (OBC), a post-training, second-order pruning and quantization framework compatible with CNNs\cite{frantar2022optimal}. It supports unstructured, block, and N:M pruning, n-bit quantization, post-optimization tuning, and post-optimization statistical correction. OBC works by breaking the problem into subproblems in layers and analyzing weights one-by-one for removal. The remaining weights are adjusted to minimize the loss of the layer. The optimal brain compression pruning algorithm ExactOBS, is a layer-wise instantiation of the optimal brain surgeon (OBS) framework. OBC establishes a pruning mask for k weights in each layer that is specified to be pruned.

\subsubsection{Torch-Pruning}
Torch-Pruning is a structured pruning approach that first analyzes the model to determine the layer-to-layer dependencies using a method called DepGraph. Using DepGraph ensures that any layers that are structurally pruned will have any dependent layers pruned as well. This ensures model cohesion and tensor dimension alignment across modified layers.

\subsubsection{TorchAO}
TorchAO is the PyTorch-provided optimization framework that includes the capability to quantize models post-training\cite{or2025torchao}. This is accomplished in a one-shot approach and requires no calibration or fine-tuning.

\subsection{Neural Architecture Search}
A comprehensive neural architecture search strategy consists of neural architecture search (NAS) with supernets and self-distillation\cite{ren2021comprehensive,pham2018efficient}. It is central to our approach for generalized AI model optimization and efficient discovery of optimal architectures. For CNNs, we have adapted NAS with supernets for efficient exploration of architectural space. It is accomplished by training a single, one shot large supernet that implicitly contains multiple sub-networks (child networks). This significantly reduces computational costs compared to traditional NAS methods, avoiding the need to train each candidate architecture after sampling to meet its performance requirements.


{One-Shot Neural Architecture Search:}
To maximize the efficacy of our NAS approach, we follow \cite{munoz2021enablingnasautomatedsupernetwork} Mu\~{n}oz et al. and provide a progressive shrinking\cite{cai_once-for-all_2020} training module to convert any CNN-based model to a weight-sharing supernet. We extend the progressive shrinking trainer by incorporating CompOFA\cite{sahni2021compofacompoundonceforallnetworks} to reduce the search space and decrease overall training time. To ensure our approach maintains compatibility with other research efforts, we provide a generic approach to encapsulate the supernet. This allows non-OFA style supernets to be used without issue. 

For the search process, we implement an evolutionary-based search that generates an initial population of sub-architectures, stored as an encoding like OFA or an extracted model object, and iterates through a series of generations, where in each generation the population is extended using crossover, mutated, and then performance predicted for the updated population. The top members of the population, based on the target constraints, are kept into the next generation. For performance prediction, the technique is decoupled from the search and supernet so that different techniques can be used, such as performance prediction models, zero-cost proxy scores, or performing a full inference session. This ensures that as new techniques are developed, they can be adapted in a plug-in-play manner into GOE.




This approach works well with larger models like LLMs and Transformers for their optimization where often the NAS approach is not practical due to vast search space associated with these models.
However this optimization approach is applicable to CNNs as well.

\subsection{Compilers}

Target edge hardware is often heterogeneous and their tensor operators and computational graphs vary from architecture to architecture. To overcome this problem, we have built a compiler module inside GOE to compile post-optimized models to to compatible to their target architecture. 

\subsubsection{Torch Compile}
PyTorch-provided compiler to speed up PyTorch models for inference. Torch compile uses Just-in-Time compilation to compiler PyTorch operations into optimized kernels.

\subsubsection{TensorRT \& TensorRT-LLM}
Nvidia's compiler and runtime engine for Nvidia hardware. TensorRT provides optional capabilities to support 2:4 pruned models and quantize to FP16 and INT8. 

\subsubsection{ONNX}
ONNX has established itself as the popular framework-independent storage format for neural networks. Popular compilers, such as TensorRT, utilize ONNX as an intermediate format between PyTorch and their compiled formats. ONNXRuntime provides the ability to execute inference of ONNX models directly in cases where further compilation is not possible.

\subsubsection{Apache TVM}
Apache TVM is a CPU and GPU agnostic compiler for increasing inference performance of neural networks. TVM supports automatic and hand-crafted compilation paths to allow quick performance gains as well as utilization of hardware knowledge to customize the compilation to a target device.

\section{Results and Discussion}
GOE targets efficient AI deployment in tactical settings where onboard computing capability and power are scarce. We decompose a global design problem into tractable subproblems whose forms depend on model scale: supernetwork-based NAS for compact CNNs, and compression for larger LLMs. A key challenge is data scarcity and sensitivity when specializing models to mission profiles.

\subsection{CNN Experimental Illustration}

\subsubsection{Structured Pruning}
For general purpose compression, structured pruning can be used to reduce the model size. Using Torch-Pruning, vision models ResNet 50 and Vision Transformer (ViT) can be compressed up to 75\%. Using a few epochs of fine tuning to recover lost accuracy, the pruned versions of the model meet, or exceed in most cases, the accuracy of the original model. Table \ref{tab:vision_base_results} shows the base accuracy and model size of ResNet 50 and ViT. Pruning these models in increments of 25\%, we can see in Table \ref{tab:torch-prune} that with only 5 epochs of fine-tuning, and in less than 20 minutes, either of these models can be compressed with pruned accuracies exceeding the uncompressed versions.

\begin{table}[!t]
\centering
\caption{Base Model Results}
\label{tab:vision_base_results}
\resizebox{0.75\columnwidth}{!}{%
\begin{tabular}{rrrr}
\toprule
Model & Accuracy (\%) & Size (Mb) \\
\midrule
ResNet 50 & 80.56 & 97.7 \\
ViT B32 & 75.82 & 336.55 \\
\bottomrule
\end{tabular}}
\end{table}

\begin{table}[!t]
\centering
\caption{Structured Pruning}
\label{tab:torch-prune}
\resizebox{\columnwidth}{!}{%
\begin{tabular}{rrrrrr}
\toprule
 & Total & & & Pruned & Pruned \\
 & Runtime & Fine-Tune & & Accuracy & Size \\
Model & (Min) & Epochs & Prune (\%) & (\%) & (Mb) \\
\midrule
ResNet 50 & 16 & 5 & 25 & 96.93 & 55 \\
ResNet 50 & 19 & 5 & 50 & 96.87 & 24.52 \\
ResNet 50 & 19 & 5 & 75 & 96.99 & 6.19 \\
ViT B32 & 18 & 5 & 25 & 96.11 & 282.51 \\
ViT B32 & 19 & 5 & 50 & 96.43 & 228.48 \\
ViT B32 & 19 & 5 & 75 & 96.27 & 174.44 \\
\bottomrule
\end{tabular}}
\end{table}

\subsubsection{Quantization Only}
When supported by target hardware, quantization can, in many cases, reduce the size of a model with minimal impact on accuracy. With the hardware support, this reduction in model size can also translate into reduction in latency. Looking at Table \ref{tab:quantization}, we see the results of applying 8-bit interger quantization using TorchAO to several CNN models. With the exception of MobileNet v3 Small, an already compressed model by design, quantization has minimal impact on the accuracy. For MobileNet models, we see over a 10x increase in throughput. For ResNet 50, that jumps to over 25x. 

\begin{table}[!t]
\centering
\caption{Quantization with TorchAO}
\label{tab:quantization}
\resizebox{\columnwidth}{!}{%
\begin{tabular}{lcccc}
\toprule
Model & Base & Base & Quant & Quant \\
 & Accuracy (\%) & Latency & Accuracy (\%) & Latency \\
\midrule
ResNet 50 & 81.0 & 0.00139 & 81.0 & 0.00005 \\
MobileNet v2 & 72.0 & 0.00059 & 72.0 & 0.00002 \\
MobileNet v3 Small & 68.0 & 0.00015 & 38.0 & 0.00001 \\
MobileNet v3 Large & 76.0 & 0.00041 & 74.0 & 0.00003 \\
\bottomrule
\end{tabular}}
\end{table}

\subsubsection{NAS - Evolutionary Search}
One of the benefits of one-shot NAS is the ability to enforce different set of constraints to target different devices at search time. In Table \ref{subnet-results-performance}, we show the results for a search targeting subnets with a maximum of 4\% accuracy loss. The search reveals the top 10 performing subnets. Not only do we achieve a 2x compression in each case, all subnets exceed the accuracy of the supernet. Looking at the results, we can also see that larger subnets do not always mean more accurate. This is the trade-off from NAS, depending on the parts of the supernet that are extracted for the subnet, the performance will vary. Looking at more than the top performing subnet based on the constraints ensure that the best model is selected for the deployment scenario. 

\begin{table}[!t]
\centering
\caption{Accuracy and Memory Size of Top-10 Sub-networks from ResNet 50 Search}
\label{subnet-results-performance}
\resizebox{\columnwidth}{!}{%
\begin{tabular}{cccccc}
\hline
Model & Number of & Compression & Accuracy & Accuracy & Memory \\
 & Parameters & Ratio & (\%) & Change & Usage (Mb) \\
\hline
ofa-rn50 & 48,105,992 & 0  & 80.14 & --  & 183.76 \\
Subnet 0 & 23,514,408 & 2.05 & 85.43 & +5.29 & 89.90 \\
Subnet 1 & 20,423,600 & 2.36 & 84.19 & +4.05 & 78.10 \\
Subnet 2 & 20,597,080 & 2.34 & 84.18 & +4.04 & 78.75 \\
Subnet 3 & 23,106,688 & 2.08 & 84.00 & +3.86 & 88.34 \\
Subnet 4 & 23,957,120 & 2.01 & 83.99 & +3.85 & 91.59 \\
Subnet 5 & 22,283,000 & 2.16 & 83.90 & +3.76 & 85.18 \\
Subnet 6 & 20,284,616 & 2.37 & 83.90 & +3.76 & 77.56 \\
Subnet 7 & 22,636,576 & 2.13 & 83.69 & +3.55 & 86.54 \\
Subnet 8 & 19,438,456 & 2.47 & 83.69 & +3.55 & 74.33 \\
Subnet 9 & 20,379,504 & 2.36 & 83.65 & +3.51 & 77.94 \\
\hline
\end{tabular}}
\end{table}

\subsubsection{NAS \& Quantization}
Our optimization pipeline is not a single method per model setup. GOE will intelligently select multiple optimization methods, when applicable, to apply to the model. For the results in Table \ref{nas_and_quant}, we apply NAS to find an optimal subnet and then apply quantization to further increase the optimization gains. In this search, we see that all candidate subnets achieve a 2x compression while exhibiting minimal accuracy loss. From the size of the subnet, quantization can achieve another 10\% drop in model size while the accuracy drop is negligible. 

\begin{table}[!t]
\centering
\caption{Accuracy and Memory Size Quantized Subnets - Supernet Accuracy: 80.14\% and Model Size: 183.77 Mb}
\label{nas_and_quant}
\resizebox{\columnwidth}{!}{%
\begin{tabular}{cccc}
\hline
Subnet & Quantized & Subnet & Quantized \\
Accuracy (\%) & Accuracy (\%) & Size (Mb) & Size (Mb) \\
\hline
78.42 & 78.34 & 93.08 & 85.26 \\
77.15 & 77.09 & 71.14 & 63.32 \\
78.07 & 78.10 & 79.73 & 73.47 \\
76.60 & 76.62 & 82.67 & 76.41 \\
77.33 & 77.36 & 77.78 & 71.52 \\
78.59 & 78.53 & 74.20 & 66.39 \\
76.97 & 76.99 & 89.46 & 83.20 \\
77.15 & 77.17 & 70.40 & 65.33 \\
77.49 & 77.42 & 61.80 & 56.73 \\
77.86 & 77.85 & 91.78 & 83.97 \\
\hline
\end{tabular}}
\end{table}

The core idea of our effort worth restating is the split by model scale. Small and mid-sized CNNs can afford full neural architecture search using supernets, because sampling and training thousands of subnetworks is computationally viable. 

The results back this up in a few places worth calling out. On ResNet50 and ViT, structured pruning at 25 to 75 percent, paired with just five epochs of fine-tuning, produced models that matched or beat the original accuracy while cutting size down to a fraction of the baseline. That's a strong signal that a lot of these networks carry more capacity than they need for their task, and a short fine-tuning pass is enough to recover whatever gets lost in pruning. Quantization told a more mixed story. TorchAO gave real latency wins, over 25x on ResNet50, with almost no accuracy hit for most models. But MobileNet v3 Small, already compact by design, took a real accuracy penalty, which is a useful reminder that quantization isn't free once a model has little redundancy left to trim.

Put together, these results support the central claim of the paper: a single optimization architecture, built around a shared set of constraints on latency, memory, and energy, can meaningfully compress both CNNs and LLMs without requiring a bespoke pipeline for each. That matters most in tactical and edge settings, where compute and power budgets are tight and the range of hardware a model might run on can't always be predicted ahead of time.

\subsection{Edge-CPU Deployment of Compressed Language Models}
\label{sec:edge-cpu}

A central promise of GOE is that its compression routes yield models that can
actually be \emph{deployed} on the constrained hardware found at the tactical
edge. We test this on the hardest such target, a device with no GPU at all, and
ask the two questions a practitioner faces before fielding a model: does the
compressed model run fast enough and fit in memory, and does it retain task
accuracy?

Two instruction-tuned language models of edge-realistic scale, Llama-3.2-1B and
Qwen2.5-1.5B, were driven through the quantization route (8-bit \texttt{Q8\_0} and
4-bit \texttt{Q4\_K\_M}) and the pruning route (structured width reduction with
recovery), then executed entirely on a CPU-only device (Intel Core Ultra~7 265U,
14 threads) using the \texttt{llama.cpp}/GGUF runtime standard for edge inference.
For each variant we measure on-disk size, single-stream decode throughput, and
accuracy on ARC-Easy, PIQA, HellaSwag, and CommonsenseQA (acc\_norm for the first
three, acc for the last).

\begin{figure}[!t]
  \centering
  \includegraphics[width=\columnwidth]{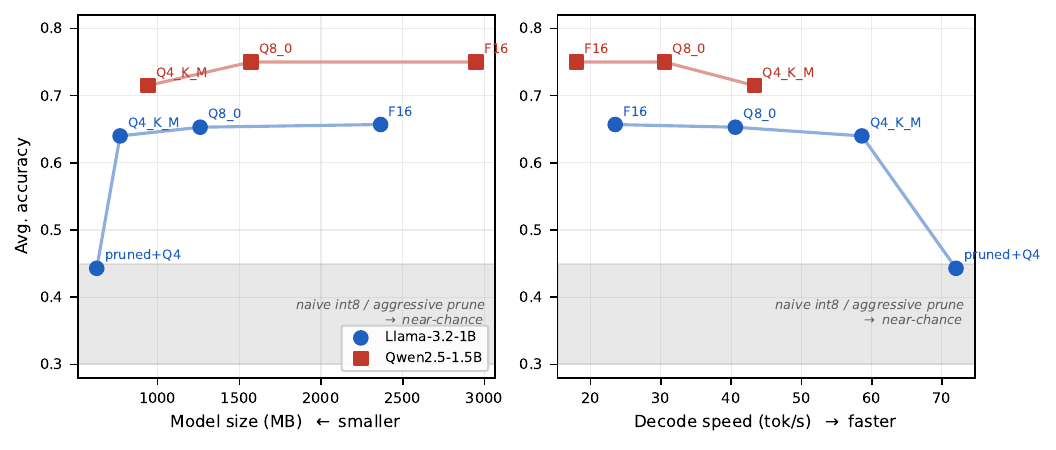}
  \caption{Edge-CPU deployment envelope on a GPU-less device (Intel Core Ultra~7,
  \texttt{llama.cpp}/GGUF). Proper GGUF quantization (\texttt{Q8\_0},
  \texttt{Q4\_K\_M}) moves each model toward smaller size (left) and higher
  throughput (right) while holding accuracy near the F16 baseline; the pruning
  route trades accuracy for further speed, and a naive dynamic-int8 scheme of the
  same 8-bit width collapses into the near-chance band.}
  \label{fig:edge-envelope}
\end{figure}

Figure~\ref{fig:edge-envelope} and Tables~\ref{tab:edge-llama}--\ref{tab:edge-qwen}
give the resulting envelope, and two findings stand out. First, \textbf{proper
edge quantization preserves accuracy while shrinking and accelerating the model.}
\texttt{Q8\_0} is nearly free: about $1.9\times$ smaller and $1.7\times$ faster
with accuracy essentially unchanged (Qwen CommonsenseQA is identical at
$0.807$). \texttt{Q4\_K\_M} delivers roughly $3\times$ smaller and
$2.5\times$ faster for a few points. Second, and more instructive for an
optimization engine, \textbf{the compression \emph{method} matters as much as the
nominal bit-width.} A naive dynamic 8-bit scheme, identical in width to
\texttt{Q8\_0}, collapses accuracy to chance on the same models through
activation-outlier error, and aggressive structured pruning degrades accuracy to
near-chance even after recovery, its speed and size gains notwithstanding. This is
exactly where an optimization engine earns its place: GOE routes each model to a
method that deploys \emph{and} works, mapping not only the deployable envelope but
the failure boundary a naive practitioner would cross unknowingly.

\begin{table}[!t]
\centering
\caption{Llama-3.2-1B on the edge CPU. \texttt{Q8}/\texttt{Q4} preserve accuracy; the pruning route trades it for speed.}
\label{tab:edge-llama}
\footnotesize
\setlength{\tabcolsep}{4pt}
\begin{tabular}{lrrcccc}
\toprule
Variant & MB & tok/s & ARC-e & PIQA & HS & CSQA \\
\midrule
F16       & 2365 & 23.5 & .620 & .760 & .627 & .620 \\
Q8\_0     & 1260 & 40.6 & .613 & .753 & .627 & .620 \\
Q4\_K\_M  &  770 & 58.6 & .613 & .760 & .600 & .587 \\
pruned+Q4 &  627 & 72.0 & .440 & .640 & .500 & .193 \\
\bottomrule
\end{tabular}
\end{table}

\begin{table}[!t]
\centering
\caption{Qwen2.5-1.5B on the edge CPU. Same pattern: quantization is the accuracy-preserving route.}
\label{tab:edge-qwen}
\footnotesize
\setlength{\tabcolsep}{4pt}
\begin{tabular}{lrrcccc}
\toprule
Variant & MB & tok/s & ARC-e & PIQA & HS & CSQA \\
\midrule
F16      & 2950 & 18.0 & .747 & .780 & .667 & .807 \\
Q8\_0    & 1570 & 30.5 & .753 & .773 & .667 & .807 \\
Q4\_K\_M &  940 & 43.3 & .747 & .747 & .633 & .733 \\
\bottomrule
\end{tabular}
\end{table}

\section{Conclusion}
We introduced the Generalized Optimization Engine (GOE), a model- and hardware-agnostic framework for deploying AI models on resource-constrained platforms. By unifying pruning, quantization, distillation, and compilation under a common optimization view, GOE formalizes deployment as a constrained or multi-objective problem over accuracy, latency, memory, and energy. We proposed decomposed formulations that adapt to model scale, enabling tractable search for both CNNs and LLMs, and described how the abstraction layer connects to existing compiler backends. We further demonstrated that GOE-compressed language models deploy and run on a GPU-less edge CPU, where proper GGUF quantization preserves task accuracy while a naive scheme of the same bit-width collapses it, so the engine's value lies in routing to a compression method that deploys \emph{and} works, not merely one that compresses.

GOE right now covers vision and language models, but tactical environments increasingly involve sensor fusion and multimodal data, and the framework needs to extend there. There's also an open question around runtime adaptivity: rather than optimizing a model once before deployment, future versions of GOE could reconfigure a model on the fly as available compute or power shifts mid-mission. Overall, our results suggest that a principled optimization view can help bridge the gap between pretrained models and real-world deployment requirements, particularly in tactical or edge settings where constraints are stringent and heterogeneous.

\section*{Acknowledgements}
This research was supported by DEVCOM Army Research Laboratory.

\bibliographystyle{IEEEtran}
\bibliography{GOE}

\end{document}